\documentclass[letterpaper, 10 pt, conference]{ieeeconf}  

\IEEEoverridecommandlockouts                              

\usepackage{amsmath,amssymb,amsfonts}
\usepackage{algorithmic}
\usepackage{graphicx}
\usepackage{textcomp}
\usepackage{xcolor}

\usepackage{tabularx} 

\def\BibTeX{{\rm B\kern-.05em{\sc i\kern-.025em b}\kern-.08em
    T\kern-.1667em\lower.7ex\hbox{E}\kern-.125emX}}

\makeatletter
\let\MYcaption\@makecaption
\makeatother

\usepackage{subcaption}
\makeatletter
\let\@makecaption\MYcaption
\makeatother

\title{User Experience Estimation in Human-Robot Interaction via Multi-Instance Learning of Multimodal Social Signals\\
}

\author{Ryo Miyoshi$^{1}$, Yuki Okafuji$^{1}$, Takuya Iwamoto$^{1}$, Junya Nakanishi$^{2}$ and Jun Baba$^{1}$ 
\thanks{$^{1}$Ryo Miyoshi, Yuki Okafuji, Takuya Iwamoto and Jun Baba are with CyberAgent, Tokyo, Japan miyoshi\_ryo@cyberagent.co.jp okafuji\_yuki\_xd@cyberagent.co.jp iwamoto\_takuya\_xa@cyberagent.co.jp baba\_jun@cyberagent.co.jp}
\thanks{$^{2}$Junya Nakanishi is with Osaka University, Osaka, Japan nakanishi@irl.sys.es.osaka-u.ac.jp}
}

\begin{document}
\maketitle

\begin{abstract}
	In recent years, the demand for social robots has grown, requiring them to adapt their behaviors based on users' states.
	Accurately assessing user experience (UX) in human-robot interaction (HRI) is crucial for achieving this adaptability.
	UX is a multi-faceted measure encompassing aspects such as sentiment and engagement, yet existing methods often focus on these individually.
	This study proposes a UX estimation method for HRI by leveraging multimodal social signals.
	We construct a UX dataset and develop a Transformer-based model that utilizes facial expressions and voice for estimation.
	Unlike conventional models that rely on momentary observations, our approach captures both short- and long-term interaction patterns using a multi-instance learning framework.
	This enables the model to capture temporal dynamics in UX, providing a more holistic representation.
	Experimental results demonstrate that our method outperforms third-party human evaluators in UX estimation.
\end{abstract}
\section{Introduction}
The declining labor force and aging population have driven the increasing deployment of social robots in service industries, including customer support and product recommendations~\cite{yamazaki2012home_robot, chen2017robotso_public_space, baba2020teleoperated, Iwamoto2022, song2022service}. 
For these robots to be effective, they must understand users' states and adjust their behavior accordingly~\cite{de2017recognizing}. 
Research has focused on estimating user states, primarily through externally observable behaviors.

People often express their psychological states through nonverbal behaviors, which play a crucial role in interactions~\cite{mandal2014nonverbal}. 
These nonverbal behaviors, collectively known as social signals~\cite{Vinciarelli_2009_socialsignal_survay}, include facial expressions, gaze, body posture, and vocal tone. 
In human-robot interaction (HRI), social signals have been widely utilized in two primary research areas: engagement assessment and sentiment analysis.

Engagement is defined as the process of initiating, maintaining, and terminating an interaction, as well as the relational state between a user and a robot~\cite{oertel2020engagement}. 
Many studies have quantified engagement as a continuous value based on observable behaviors such as gaze direction and posture~\cite{anzalone2015gaze_engagement_est, Rudovic_2019_CVPR_Workshops, del2020you, del2022learning}. 
While these engagement estimation models have been applied to robot behavior design, they primarily capture the level of interaction rather than the user's subjective evaluation of the experience.

Sentiment analysis, on the other hand, attempts to classify a user's internal state as positive, neutral, or negative based on multimodal inputs~\cite{soleymani2017sentiment_survey}. 
This approach has been widely used in affective computing and HRI to design behaviors that promote positive emotional responses~\cite{clavel2015sentiment_analysis, tomimasu2016assessment, tavabi2019multimodal, yao2020morse, katada2022hazumi}. 
However, sentiment analysis provides only a simplified representation of user experience, making it challenging to capture the nuanced and multi-faceted aspects of interactions.

In contrast, user experience (UX) is a broader and more comprehensive concept that encompasses various aspects of an interaction~\cite{UX_book}. 
UX evaluations are commonly conducted using multi-dimensional scales, such as the User Experience Questionnaire (UEQ)~\cite{schrepp2015ueq}, which assesses aspects like attractiveness, efficiency, and novelty.

While UX, engagement, and sentiment are distinct concepts, they are closely interrelated and influence one another. 
Engagement reflects the extent to which a user is involved in an interaction, while sentiment captures momentary emotional states. 
UX, in contrast, is a higher-level, multi-dimensional construct that integrates both engagement and sentiment, as well as other cognitive and affective responses. 
For example, a user may be highly engaged but still have a negative UX if the interaction feels frustrating. 
Conversely, positive sentiment alone does not necessarily translate to a satisfying UX if the interaction lacks efficiency or usefulness~\cite{okafuji2022robot_in_mall}. 

Since UX directly affects user satisfaction and the long-term adoption of robotic services, estimating UX in HRI is crucial for improving user-centered robot behaviors~\cite{shourmasti2021UX_SR, forgas2022UX_SR_eval}. 
However, despite the growing interest in HRI, few studies have focused on automatically evaluating UX in human-robot interactions.  
Most existing research relies on post-interaction surveys, which are subjective and time-consuming, making large-scale analysis difficult.  
Moreover, while engagement and sentiment have been extensively studied in HRI, there is no established approach for systematically modeling and evaluating UX in an automated manner.

To address this limitation, this study proposes an approach to automating UX evaluation in HRI by leveraging multimodal social signals.  
Specifically, we introduce a multi-instance learning framework that captures both short-term and long-term interaction patterns to assess UX.  
Unlike conventional engagement or sentiment models that are based on momentary observations, our method analyzes sequential patterns to construct a more comprehensive representation of UX.  
While large-scale vision-language models (VLMs) excel in many tasks, they struggle with video data, making it difficult to capture both short-term and long-term features in interactions.  
To overcome this, we evaluate our proposed method using data from human-robot interactions.  

The main contributions of this study are as follows:
\begin{itemize}
    \item We propose a Transformer-based multi-instance learning model for UX estimation in human-robot interaction, capturing both short- and long-term patterns.
    \item We evaluate the proposed method by comparing its accuracy against multiple baselines, including third-party human evaluators and unimodal models.
    \item We analyze the effectiveness of facial expressions and voice in UX estimation, demonstrating the benefits of a multimodal approach.
\end{itemize}

\section{Related Work}
\subsection{Assessment of Engagement}
Engagement is defined as the process of initiating, maintaining, and terminating an interaction, as well as the relational state between a user and an entity~\cite{oertel2020engagement}. 
In HRI, engagement is often measured as the degree to which a user is involved in an interaction with a robot. 
Many studies have proposed methods to estimate and enhance engagement by analyzing observable behavioral cues such as gaze direction, body posture, and facial expressions~\cite{anzalone2015gaze_engagement_est, Rudovic_2019_CVPR_Workshops, del2020you, del2022learning}.

Engagement is typically quantified as a continuous value ranging from 0 to 1, representing an externally observable state in HRI.
For instance, Salvatore et al.~\cite{anzalone2015gaze_engagement_est} estimated engagement using user gaze and gestures.
Rudovic et al.~\cite{Rudovic_2019_CVPR_Workshops} developed a deep-learning-based approach for engagement estimation in child–robot interactions.
Del et al.~\cite{del2020you} introduced an engagement assessment method using first-person robot vision, while Del et al.~\cite{del2022learning} applied reinforcement learning to maximize engagement in tour-guide robots.

Although engagement estimation provides valuable insights into user-robot interactions, it has inherent limitations. 
Engagement is an \textit{externally observable state} rather than a direct measure of user perception.
Annotations are typically provided by third-party observers, which may not necessarily align with the user's subjective evaluation~\cite{truong2012speech}. 
For instance, a user who maintains eye contact with a robot might be classified as highly engaged, even if they feel uncomfortable or frustrated.
This suggests that engagement alone is insufficient to fully capture the overall quality of an interaction.

\subsection{Sentiment Analysis}
Sentiment analysis estimates a user's psychological state by classifying their emotional response into discrete categories such as positive, neutral, or negative~\cite{soleymani2017sentiment_survey}. 
This method has been widely utilized in affective computing and HRI for developing adaptive robot behaviors~\cite{clavel2015sentiment_analysis, tomimasu2016assessment, tavabi2019multimodal, yao2020morse, katada2022hazumi}.

Several studies have explored multimodal sentiment analysis. 
Tomimasu et al.~\cite{tomimasu2016assessment} proposed an emotion estimation method using visual and vocal cues during user–system interactions.
Yao et al.~\cite{yao2020morse} developed a multimodal sentiment analysis framework trained on real-world interaction datasets.
Katada et al.~\cite{katada2022hazumi} investigated the effectiveness of physiological and observable signals in estimating subjective and third-party sentiment ratings.

\begin{table}[t]
    \caption{Robot Behavior That Changes the User Experience}
    \centering
    \begin{tabularx}{\linewidth}{XX}
        \hline
        Robot behavior that leads to \par negative user experience & Robot behavior that leads to \par positive user experience \\ \hline \hline
        \begin{minipage}{40truemm} 
            \begin{itemize}
                \item Interrupt someone speaking
                \item Dialogue delay
                \item Repeat the same question back
                \item Hear incorrectly
                \item Duplicate description
            \end{itemize}
        \end{minipage}
        &
        \begin{minipage}{40truemm} 
            \begin{itemize}
                \item Pop performance with music and dance
                \item Dialogue with the right timing
                \item Speech with good voice quality
            \end{itemize}
        \end{minipage} \\ \hline
    \end{tabularx} 
    \vspace{-1em}
    \label{table:robot_action}
\end{table}

While sentiment analysis provides useful insights into momentary user emotions, it has inherent limitations when applied to HRI.
Classifying emotions into broad categories (e.g., positive, neutral, negative) oversimplifies the complexity of user experience.
For example, a user may feel excitement but also frustration due to an inefficient interaction.
Additionally, sentiment analysis does not account for factors such as usability or task effectiveness, which are essential for evaluating the overall quality of interaction.

\subsection{User Experience in Human–Robot Interaction}
User experience (UX) is a broader concept that encompasses various aspects of an interaction, including user perceptions, emotional responses, and cognitive evaluations~\cite{UX_book}. 
According to ISO 9241-210, UX is defined as ``a person's perceptions and responses that result from the use and/or anticipated use of a product, system, or service.''
Unlike engagement, which focuses on the degree of involvement, and sentiment, which captures emotional states, UX integrates these aspects along with usability and interaction effectiveness.

In HRI, UX is commonly assessed through post-interaction questionnaires.
Shourmasti et al.~\cite{shourmasti2021UX_SR} explored UX evaluation challenges in social robots, emphasizing its role in improving behavioral design.
Forgas et al.~\cite{forgas2022UX_SR_eval} analyzed key UX factors in social robot services and highlighted the importance of reliable evaluation methods.

While engagement and sentiment are important components of UX, they do not fully define it.
For example, a user may be highly engaged in an interaction and exhibit positive emotions, but if the interaction lacks efficiency or responsiveness, their overall UX may still be negative.
Thus, UX offers a more holistic assessment of interaction quality by integrating engagement, sentiment, and other user-centered factors.

\section{Collection of Data on User Experience in Human-Robot Interaction}
In this study, to estimate UX based on social signals in HRI, we created a dataset containing UX ratings and user interactions in HRI. 
For the HRI experiments, two types of robots were used, and interactions were designed to elicit different experiences across multiple scenarios. 
To evaluate UX, we collected user experience ratings through questionnaires. 
This study was approved by the Research Ethics Committee of Osaka University (Reference number: 31-1-2).

The experiment was conducted with 22 participants. 
Most participants had limited prior experience interacting with robots, and this was their first time engaging with the specific robots used in this study. 
Among the participants, 22 were male, and one was female. 
The average age of the participants was 21.1 years. 
Each participant received 1,000 JPY per hour for their participation.

\begin{figure}[t]
    \begin{minipage}[ht]{0.48\columnwidth}
        \centering
        \includegraphics[width=\columnwidth]{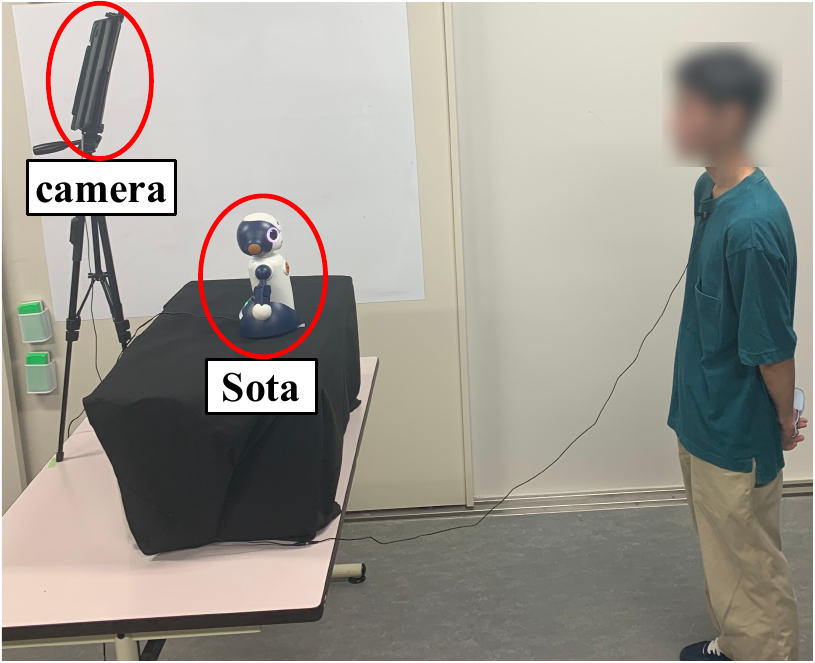}
        \subcaption{Scene of interaction with Sota}
    \end{minipage}
    \hspace{0.02\columnwidth} 
    \begin{minipage}[ht]{0.48\columnwidth}
        \centering
        \includegraphics[width=\columnwidth]{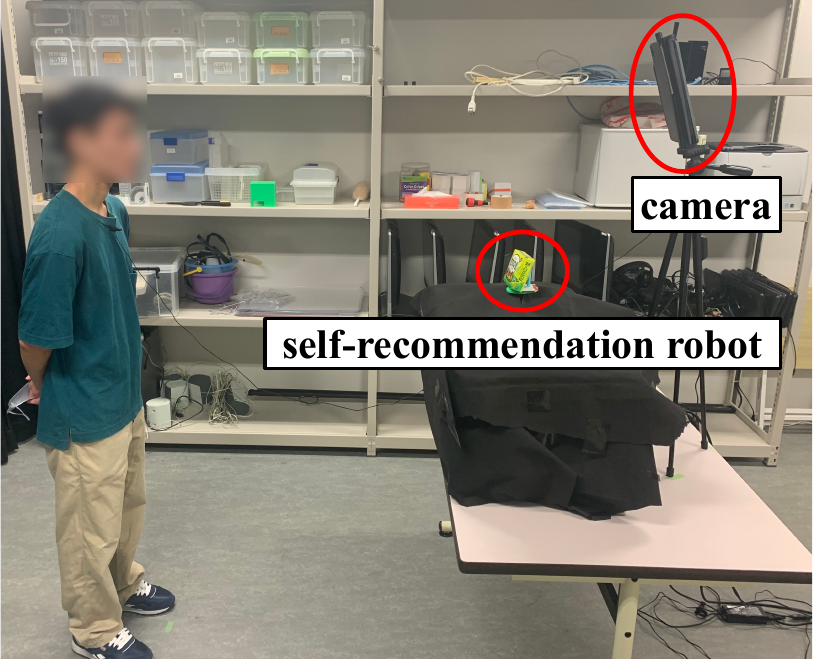}
        \subcaption{Scene of interaction with self-recommendation robot}
    \end{minipage}
    \caption{Scenes from the interaction experiment with robots}
    \label{fig:summary_experiment}
\end{figure}

\begin{figure}[t]
    \centering
    \includegraphics[width=0.35\linewidth]{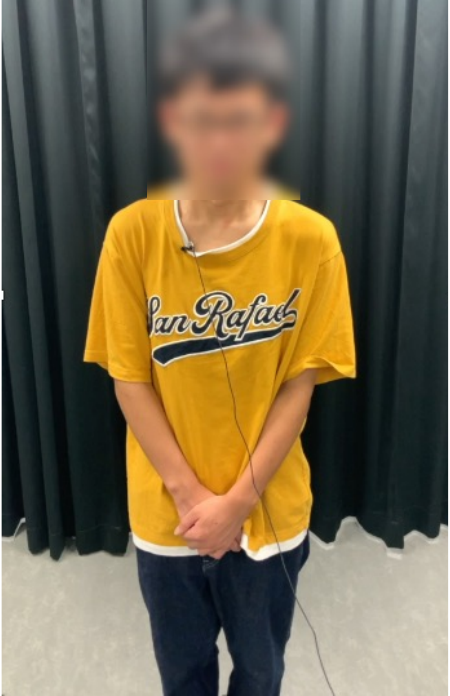}
    \caption{Example of participant images captured by the camera}
    \vspace{-1em}
    \label{fig:summary_camera}
\end{figure}

\begin{figure}[t]
    \centering
    \includegraphics[width=1\linewidth]{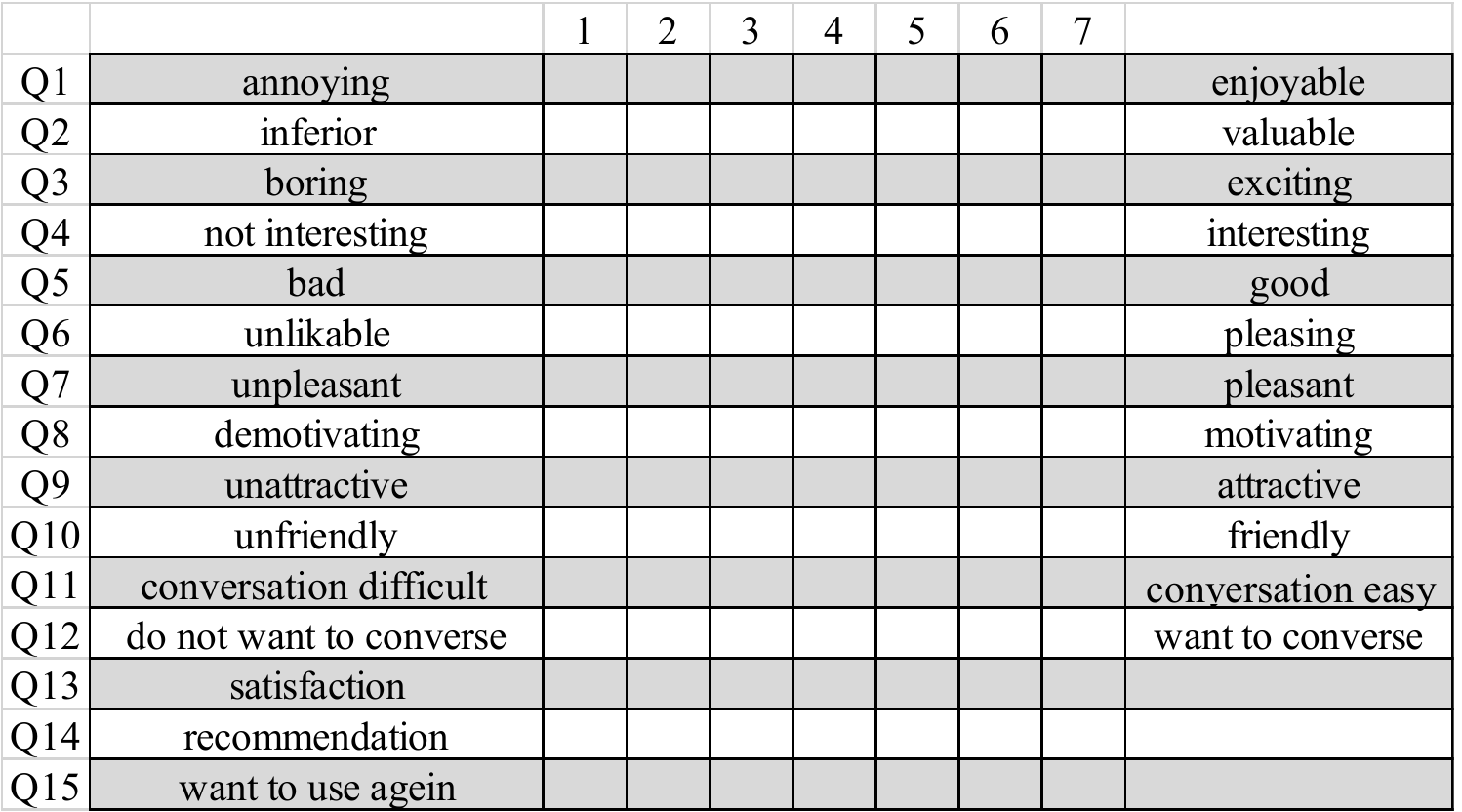}
    \caption{Questionnaire used in this experiment}
    \vspace{-1em}
    \label{fig:questionnaire}
\end{figure}

\subsection{Interaction Settings}
\label{sec:inter_setting}
Two types of robots were used in this experiment: Sota (``Sota'' developed by Vstone Co., Ltd.), a small humanoid robot, and a self-recommendation robot~\cite{Iwamoto2022}. 
During the experiment, participants interacted with the robots in predefined scenarios, such as discussing food preferences, hobbies, and new product introductions. 
To elicit different user experiences, we designed scenarios incorporating robot behaviors intended to create either positive or negative user experiences. 
Table~\ref{table:robot_action} summarizes the robot behaviors associated with positive and negative experiences. 

For interactions with Sota, five scenarios were designed to induce a negative user experience. 
Conversely, for the self-recommendation robot, five scenarios were created to foster a positive user experience. 
In total, 10 scenarios were prepared, and they were assigned to each participant in a random order. 
We employed two types of robots to facilitate the elicitation of both positive and negative experiences. 
Since this study aimed to estimate UX based on social signals, the difference between the two robots was expected to have minimal impact on the overall user experience.

Fig.~\ref{fig:summary_experiment} illustrates the experimental setup. 
The experiment was conducted in a controlled laboratory environment. 
The robots were operated using the Wizard of Oz method, in which an operator remotely controlled the robots. 
The distance between the robot and the participants was set to 100 cm, and the robots were positioned at a height of 105 cm above the floor. 
This height ensured that participants' faces remained level, allowing for accurate capture of their gaze and facial expressions. 

A camera was positioned behind the robot at a height of 155 cm above the floor to record videos of the participants. 
This placement was chosen to capture the participants' full bodies, as social signals are expressed through entire-body movements~\cite{Vinciarelli_2009_socialsignal_survay}. 
Additionally, each participant's voice was recorded using a pin microphone. 
Fig.~\ref{fig:summary_camera} presents an example of an image captured by the camera, illustrating the participant's appearance.

\subsection{Evaluation of User Experience}
In this experiment, participants completed a UX evaluation questionnaire after each interaction. 
The questionnaire was developed based on the User Experience Questionnaire (UEQ)~\cite{schrepp2015ueq}, which is commonly used to assess the quality of interactive products. 
The UEQ consists of 26 items across six dimensions: Attractiveness, Perspicuity, Efficiency, Dependability, Stimulation, and Novelty. 
Attractiveness reflects the overall impression of the product. 
Perspicuity, Efficiency, and Dependability assess its pragmatic quality, while Stimulation and Novelty evaluate its hedonic quality.

Since not all UEQ items were applicable to UX evaluation in this study, we selected only those related to Attractiveness and Stimulation, which are considered relevant for assessing user experience during interactions. 
Additionally, UX evaluation in social robot interactions typically includes aspects of interactivity and satisfaction~\cite{baba2020teleoperated}. 
Therefore, we incorporated direct evaluation items for interaction quality and satisfaction derived from the interaction. 
Fig.~\ref{fig:questionnaire} presents the questionnaire used in this experiment.

The final questionnaire consisted of 15 items: 10 items (Q1–Q10) from the UEQ, two items assessing interaction quality (Q11 and Q12), and three items evaluating satisfaction (Q13–Q15). 
Items Q1–Q12 were rated using the semantic differential method, while items Q13–Q15 were assessed using a Likert scale. 
This questionnaire measured UX using a discrete scale ranging from 1 to 7.

\begin{figure*}[t]
    \centering
    \includegraphics[width=\textwidth]{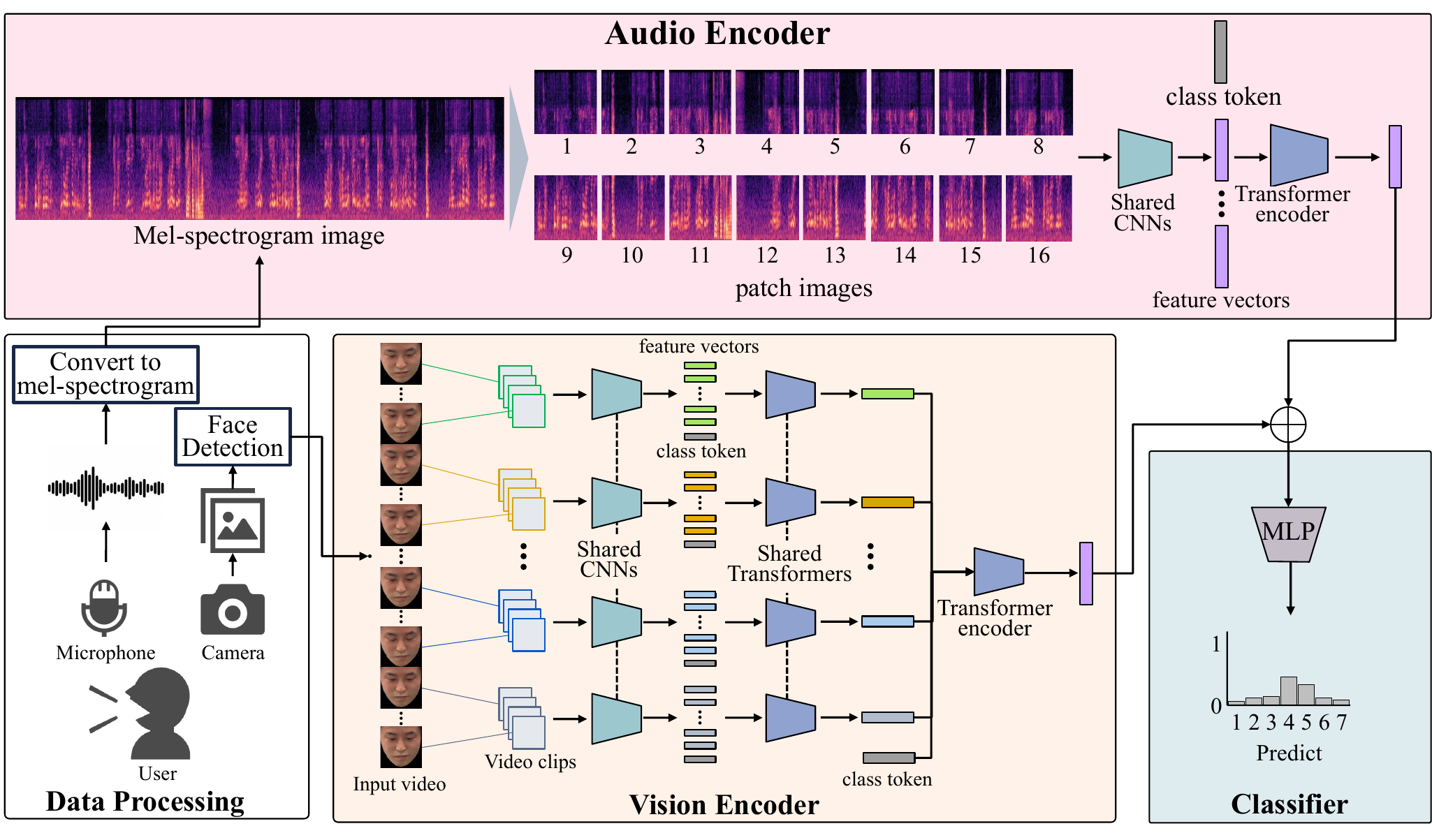}
    \captionsetup{skip=1pt}
    \caption{Overview of the proposed UX estimation model}
    \label{fig:proposed}
    \vspace{-1em}
\end{figure*}

Each of the 22 participants interacted with both robots across a total of ten scenarios, five per robot. 
After completing each interaction, participants filled out the questionnaire. 
A total of 220 data points were collected. 
The minimum and maximum interaction durations were 30 and 120 seconds, respectively, with an average of approximately five conversational turns per interaction.

\section{Proposed UX Estimation Method}
In this study, we propose a UX estimation method that utilizes social signals—observable cues reflecting a user's psychological state—as input. 
Social signals include posture, gaze, head movements, facial expressions, and voice~\cite{Vinciarelli_2009_socialsignal_survay}. 
Although all these modalities were recorded, posture, gaze, and head movements exhibited minimal variation due to the controlled experimental setup. 
Additionally, since the interactions were casual conversations rather than task-oriented dialogues, textual content was not utilized. 
Therefore, we focused on facial expressions and voice for UX estimation.

Fig.~\ref{fig:proposed} presents an overview of the proposed method. 
UX is not evaluated based on momentary impressions but rather through a sequence of interactions. 
Moreover, as the stimuli influencing UX constitute only a subset of the entire interaction sequence, it is essential to consider both short-term and long-term features. 
To address this, we propose a model incorporating the concept of multiple instance learning. 
This approach enables the extraction of short-term features from individual instances while capturing long-term features from the aggregated set of instances.

\subsubsection{Data Processing}
User data, including video and audio, are recorded using a camera and a microphone during interactions. 
For video data, a face detection algorithm is applied to each frame, and the detected face region is cropped.
Simultaneously, the recorded audio is transformed into a mel-spectrogram representation.
The processed mel-spectrogram images serve as input for the audio encoder, while the facial images are fed into the vision encoder.

\subsubsection{Audio Encoder}
The audio encoder processes the mel-spectrogram representation of the recorded speech to extract relevant acoustic features for UX estimation.
Previous studies in speech emotion recognition have demonstrated high accuracy using mel-spectrogram images as input to convolutional neural networks (CNNs)~\cite{Dossou_2021_ICCV}. 
Additionally, an audio classification method utilizing patch-based mel-spectrogram inputs for a Transformer model has been proposed~\cite{AST_interspeech_2021}.
Based on these findings, our model employs a combination of CNN and Transformer architectures for feature extraction.

In the audio encoder, the mel-spectrogram image generated from the user's voice is first divided into a series of patch images $X_a \in \mathbb{R}^{M_a \times h_a \times w_a}$.
Each patch represents a local segment of the mel-spectrogram image, where $P$ is the number of patches, and $h_a$ and $w_a$ are the height and width of each patch, respectively.

Then the patches are fed into a CNN-based feature extractor to extract local spectral features $F_a$.
\begin{equation}
    F_a = \text{CNN}(X_a)
\end{equation}
where $F_a \in \mathbb{R}^{N_a \times d_a}$ is the extracted feature vector from the CNN, and $d_a$ is the dimension of the feature vector.
The CNN used for feature extraction shares its weights across all patches, ensuring a consistent feature representation.
In addition, CNN is initialized using weights pre-trained on ImageNet~\cite{ImageNet_CVPR_2009}.

The feature vectors extracted from all patches are fed into a Transformer encoder, which captures sequential dependencies and long-range interactions within the speech signal.
The Transformer processes these feature vectors $F_a$, along with a $CLS$ token that aggregates the extracted information.
As a result, a global feature representation of the audio signal, denoted as $Z_a$, is generated.
\begin{equation}
    Z_a = \text{Transformer}([F_a, CLS])
\end{equation}
where $Z_a \in \mathbb{R}^{d}$ is the global feature representation of the audio signal and $CLS \in \mathbb{R}^{d}$ is the class token.

To generate a unified representation of the input audio, the Transformer encoder processes the feature vectors and outputs a single global feature representation.

\subsubsection{Vision Encoder}
The vision encoder processes facial images extracted from the recorded video to capture relevant visual features for UX estimation. In facial expression recognition, methods that split a video into shorter video clips are commonly used when processing long video sequences~\cite{snippet_FER_2023_PR}. Additionally, Transformer-based architectures have demonstrated high performance in modeling temporal dependencies for facial expression analysis~\cite{Former-DFER_ACMMM_2021, snippet_FER_2023_PR, Logo-Former_ICASSP_2023}.
Based on these findings, our model employs a combination of CNN and Transformer architectures for feature extraction.
Furthermore, the proposed method adopts two-stage multi-instance learning to extract more local and global feature values.

In the vision encoder, the input video is first divided into a set of video clips, which are then represented as image frames.
Each frame is processed by a CNN-based feature extractor to extract spatial features $C_i$:
\begin{equation}
    C_i = \text{CNN}(X_{v_i})
\end{equation}
where $X_v \in \mathbb{R}^{T \times h_v \times w_v}$ represents the set of video frames, $T$ is the number of frames, and $h_v$ and $w_v$ are the height and width of each frame, respectively.
The index of the video clip is denoted as $i$, where $i \in \{1, \dots, M\}$.
The extracted feature vectors, $C \in \mathbb{R}^{N_v \times d_v}$, are then fed into a Transformer encoder to model temporal dependencies across frames.
Here, the CNN used for feature extraction shares its weights across all video frames, ensuring a consistent feature representation.
The CNN is initialized with weights pre-trained on the FER-Plus facial expression dataset~\cite{FER_plus}.

The first Transformer encoder aggregates frame level features to capture temporal dependencies within the video clip.
The feature vectors $C_i$ extracted from individual frames are processed by the Transformer encoder 
The feature vectors $C_i$ are processed along with a $CLS$ token that aggregates the extracted information, denoted as $F_{v_i}$:
\begin{equation}
    F_{v_i} = \text{Transformer}([F_{v_i}, CLS_i])
\end{equation}
where $F_v \in \mathbb{R}^{d}$ is the feature vector of facial expressions for video clip $i$, and $CLS_i \in \mathbb{R}^{d}$ is the class token.

The second Transformer encoder aggregates the feature vectors $F_{v_i}$ across all video clips to capture long-term dependencies.
The feature vectors $F_{v_i}$ are processed by the Transformer encoder, which outputs a single global feature representation $Z_v$:
\begin{equation}
    Z_v = \text{Transformer}([F_{v_1}, F_{v_2}, ..., F_{v_M}, CLS])
\end{equation}
where $Z_v \in \mathbb{R}^{d}$ is the global feature representation of the facial expressions in the video.

\subsubsection{Classifier}
After feature extraction, the obtained feature vectors from the audio and vision encoders are fused to form a unified multimodal representation.
This fused representation is processed by a shared Transformer module to capture cross-modal dependencies.
Finally, a multi-layer perceptron (MLP) classifier predicts the UX score based on the combined feature embeddings.
\begin{equation}
    \hat{y} = \text{MLP}([Z_a, Z_v])
\end{equation}
where $\hat{y} \in \{1, 2, ..., 7\}$ is the predicted UX score, and $[Z_a, Z_v]$ is the concatenated feature vector from the audio and vision encoders.

The loss function of the proposed method is defined as cross-entropy loss:
\begin{equation}
    \mathcal{L} = -\sum_{i=1}^{N} y_i \log(\hat{y}_i)
\end{equation}
where $y_i$ is the ground truth label, and $\hat{y}_i$ is the predicted label.

\begin{figure*}[ht]
    \centering
    \includegraphics[width=1\linewidth]{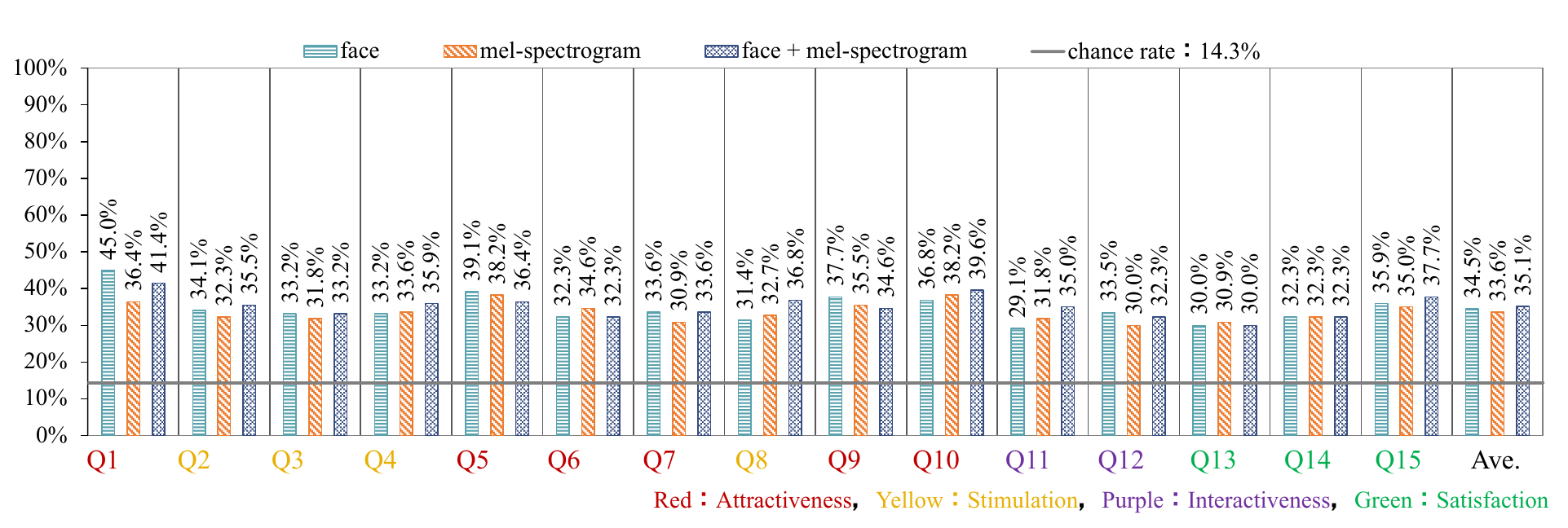}
    \captionsetup{skip=1pt}
    \caption{Acc.7 of each model}
    \vspace{-1em}
    \label{fig:acc_model}
\end{figure*}

\begin{figure*}[t]
    \centering
    \includegraphics[width=1\linewidth]{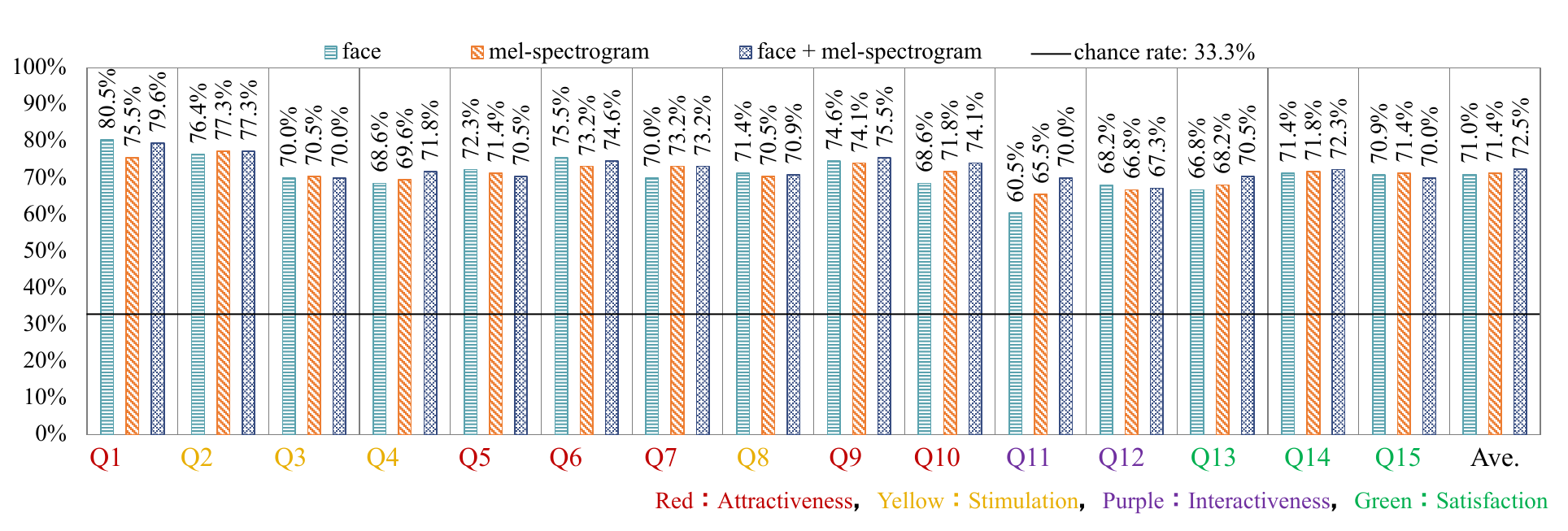}
    \captionsetup{skip=1pt}
    \caption{Acc.3 of each model}
    \vspace{-1em}
    \label{fig:abs_model}
\end{figure*}

\section{Experiments}
\label{sec:ux_extimate}

\subsubsection{Experimental Settings}
The estimation accuracy of each model was evaluated using subject-independent $k$-fold cross-validation, with $k$ set to 4. 
Additionally, the models were assessed using two evaluation metrics: Acc.7 and Acc.3.  

Acc.7 represents the accuracy of predictions for questionnaire items scored on a seven-point scale.  
Acc.3 is a quantized version of Acc.7, where the original seven-scale responses are grouped into three categories:  
responses 1 and 2 are categorized as the negative scale (0), responses 3, 4, and 5 as the neutral scale (1), and responses 6 and 7 as the positive scale (2).  
The estimation target was the response to a single questionnaire item.

The model training conditions were as follows: the batch size was set to 4, the number of video clips and the number of patch images was set to 16, and each video clip contained 16 frames.  
Facial images were extracted using OpenFace~\cite{Openface}, with a resolution of $112 \times 112$ pixels.  
The proposed model was implemented using PyTorch and trained on an NVIDIA RTX 3090 GPU.

\subsection{Ablation Study}
To evaluate the effectiveness of the proposed multimodal model, we compared three models: a model that used only facial images as input (vision-only encoder), a model that used only audio as input (audio-only encoder), and a multimodal model that incorporated both inputs. 

The results of this experiment are presented in Fig.~\ref{fig:acc_model} and Fig.~\ref{fig:abs_model}. 
As shown in Fig.~\ref{fig:acc_model}, the average Acc.7 across all questionnaire items was 34.5\% for the vision-only model, 33.6\% for the audio-only model, and 35.1\% for the multimodal model using both facial images and voice. 
Since the chance rate for Acc.7 was 14.3\%, each model performed significantly better than chance.

Regarding attractiveness, Q1 (``annoying - enjoyable''), Q5 (``bad - good''), and Q10 (``unfriendly - friendly'') achieved higher accuracy compared to other items within the same scale. 
In particular, Acc.7 for the vision-only model was higher for Q1 (``annoying - enjoyable'') and Q5 (``bad - good''), suggesting that internal states are reflected in facial expression changes. 
Additionally, the accuracy of Q8 (``demotivating - motivating'') and Q11 (``conversation difficult - conversation easy'') improved when both facial images and voice were combined, compared to unimodal models.

The estimation accuracy for many items related to Interactivity and Satisfaction was approximately 30.0\%, indicating that these aspects were more challenging to estimate than other scales. 
As shown in Fig.~\ref{fig:abs_model}, when the metric is Acc.3, the chance rate is 33.3\%, and the mean estimation accuracy exceeded 70.0\% for all models, confirming that UX estimation is feasible. 
Fig.~\ref{fig:abs_model} also demonstrates that all models achieved an average Acc.3 of at least 70.0\%, with the proposed model outperforming the other models on most questionnaire items.

\begin{figure*}[t]
    \centering
    \includegraphics[width=1\linewidth]{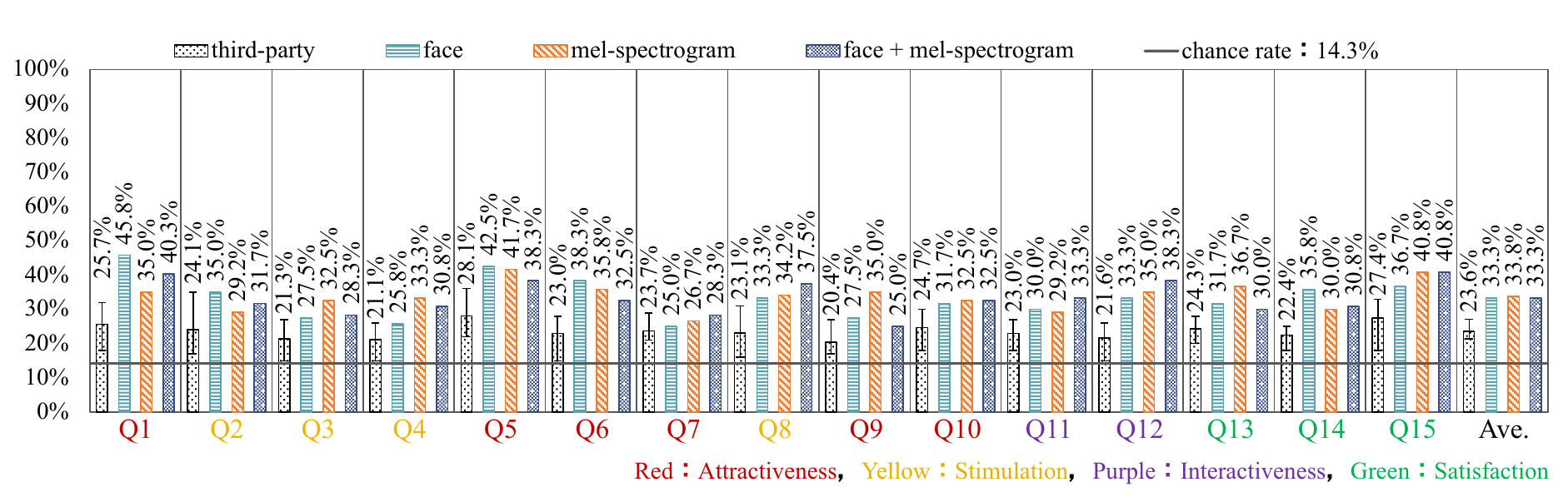}
    \captionsetup{skip=1pt}
    \caption{Acc.7 of each model and a third-party model}
    \vspace{-1em}
    \label{fig:acc_human}
\end{figure*}

\begin{figure*}[t]
    \centering
    \includegraphics[width=1\linewidth]{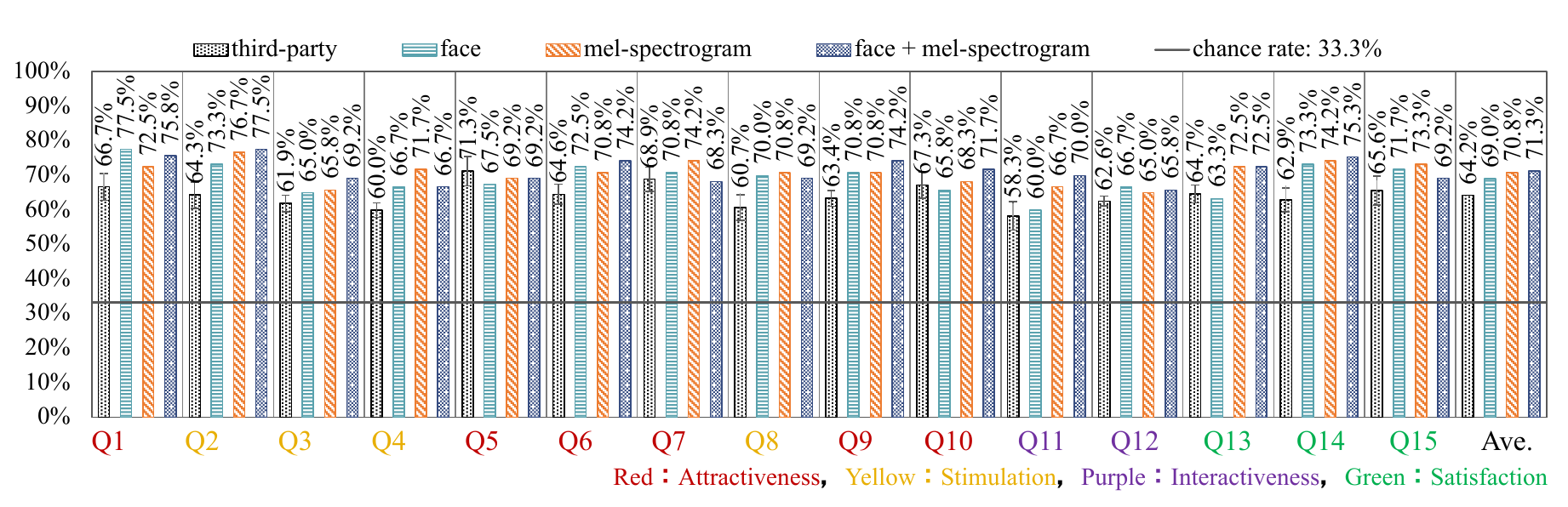}
    \captionsetup{skip=1pt}
    \caption{Acc.3 of each model and a third-party estimation}
    \label{fig:abs_human}
\end{figure*}

\begin{figure*}[ht]
    \begin{minipage}[b]{0.23\textwidth}
        \centering
        \includegraphics[width=\textwidth]{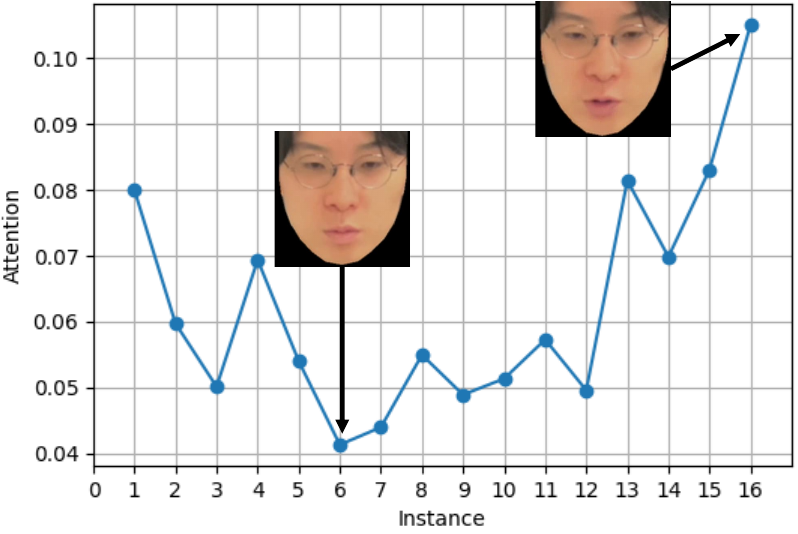}
        \subcaption{Attention of the only vision encoder model}
    \end{minipage}
    \hspace{0.01\textwidth} 
    \begin{minipage}[b]{0.23\textwidth}
        \centering
        \includegraphics[width=\textwidth]{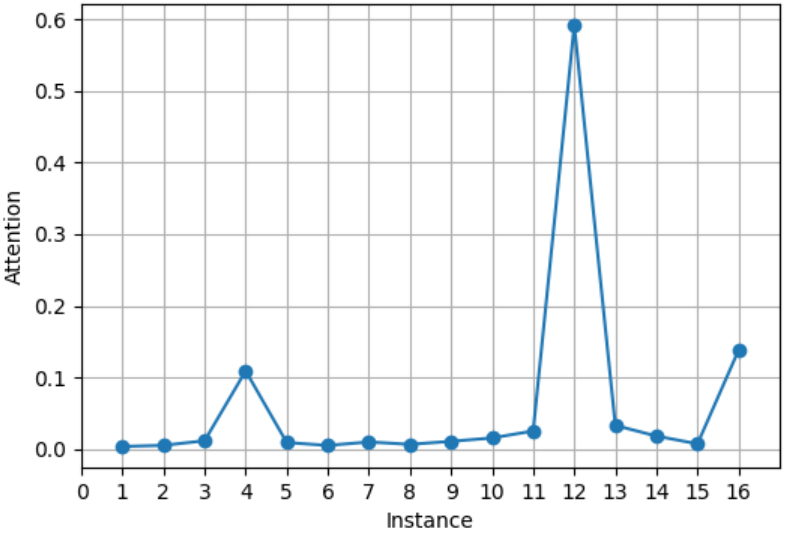}
        \subcaption{Attention of the only audio encoder model}
    \end{minipage}
    \hspace{0.01\textwidth} 
    \begin{minipage}[b]{0.23\textwidth}
        \centering
        \includegraphics[width=\textwidth]{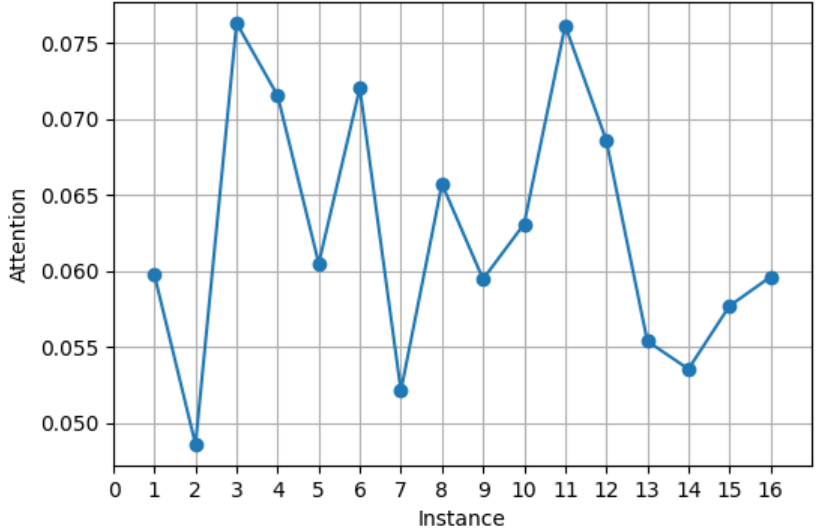}
        \subcaption{Attention on face for the proposed model}
    \end{minipage}
    \hspace{0.01\textwidth} 
    \begin{minipage}[b]{0.23\textwidth}
        \centering
        \includegraphics[width=\textwidth]{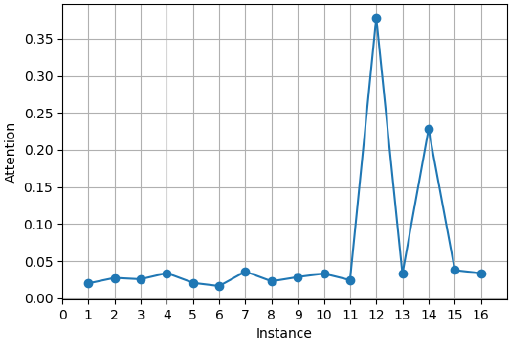}
        \subcaption{Attention on voice for the proposed model}
    \end{minipage}
    \caption{Examples of attention obtained from each model}
    \vspace{-1em}
    \label{fig:atts}
\end{figure*}

\subsection{Comparison of Each Model and Third-Party Estimation}
Previous studies have evaluated model performance by comparing it with third-party estimations~\cite{borji2014human_vs_ml}. 
Following this approach, we conducted a third-party UX estimation to objectively assess the proposed model's performance.

This experiment used data from 12 of the 22 participants who granted permission for its use. 
Evaluators were provided with video recordings, including audio, and tasked with estimating participants' questionnaire responses. 
The procedure was as follows:  
First, each evaluator watched videos of two participants as training data to familiarize themselves with their questionnaire responses.  
Next, they watched videos of the remaining 10 participants and estimated their questionnaire values.  
Seven independent evaluators with no prior knowledge of the study participated to ensure unbiased assessments.  
The proposed model's performance was averaged over the 12 participants.

Fig.~\ref{fig:acc_human} and Fig.~\ref{fig:abs_human} present the results. 
Error bars in the third-party evaluation graph indicate the maximum and minimum accuracy values, representing the best and worst evaluators.  
The chance rates for Acc.7 and Acc.3 were 14.3\% and 33.3\%, respectively.  
For Acc.7, all proposed models achieved approximately 33.0\%, while third-party estimation was around 23.0\%.  
The proposed method outperformed the best third-party evaluator for many items.  

For Acc.3, third-party estimation reached 64.0\%, while the proposed model achieved 70.0\%.  
Although the number of items where the best third-party evaluator surpassed the proposed model increased compared to Acc.7, the proposed method still performed as well as or better than the best evaluator in more than half of the items.

\subsection{Analysis of Attention for Each Model}
For each time interval, we analyzed where each model focused during the interaction to estimate UX. The analysis first computed the attention of the Transformer encoder, which aggregates features from video clips and patch images using attention rollout~\cite{abnar2020att_rollout}.
The corresponding video and voice segments were then manually examined.

Fig.~\ref{fig:atts} presents an example of attention visualization. In the facial image attention graph, the horizontal axis represents video clips, with instances closer to the right corresponding to later interaction stages.
In the voice attention graph, the horizontal axis aligns with the top (1–8) and bottom (9–16) rows of patch images in Fig.~\ref{fig:proposed}-(b).

The analysis revealed that attention to facial images was distributed across multiple instances, suggesting UX estimation occurred throughout the interaction.
In contrast, attention to voice was concentrated on specific instances despite multiple utterances. Facial image attention was higher for clips featuring expressive changes, such as increased mouth corner elevation and surprise (Fig.~\ref{fig:atts}-(a)).
Similarly, voice attention peaked when tonal variations were detected. These findings highlight the importance of capturing dynamic changes in the user's state.

\section{Conclusion}
In this study, we focused on UX estimation in HRI as a comprehensive measure that encompasses multiple aspects of user experience.
We constructed a dataset and proposed UX estimation method that utilize social signals, specifically facial expressions and voice, as input.

Experimental results demonstrated that the multimodal model incorporating higher performance than models using a single modality.
To objectively assess the performance of the proposed method, we compared its performance with third-party UX estimations.
The results confirmed that the proposed approach outperforms third-party evaluations, indicating its effectiveness in UX estimation.

Although the results are promising, this study was conducted in a controlled environment and may not fully reflect real-world user behavior. Future work includes validating the method in real-world settings, incorporating UX estimation into robot behavior strategies, and developing real-time estimation frameworks for feedback-based interactions.

{
    \small
    \bibliographystyle{IEEEtran}
    \bibliography{IEEEabrv,bibtex}
}

\end{document}